\documentclass{article}

\usepackage[preprint]{neurips_2023}

\usepackage[utf8]{inputenc} 
\usepackage[T1]{fontenc}    
\usepackage{hyperref}       
\usepackage{url}            
\usepackage{booktabs}       
\usepackage{amsfonts}       
\usepackage{nicefrac}       
\usepackage{microtype}      
\usepackage{xcolor}         
\usepackage{graphicx}
\usepackage{amsmath}
\usepackage{appendix}
\usepackage{mathtools}
\usepackage[capitalize]{cleveref}
\usepackage{algorithm2e}
\usepackage{multirow}
\crefname{section}{Sec.}{Secs.}
\Crefname{section}{Section}{Sections}
\Crefname{table}{Table}{Tables}
\crefname{table}{Tab.}{Tabs.}
\crefname{algocf}{Alg.}{Algs.}
\Crefname{algocf}{Algorithm}{Algorithms}

\newcommand{\cp}{c_{p}(\bfx)}
\newcommand{\cid}{c_{id}(\bfx)}
\newcommand{\bfx}{\mathbf{x}}

\newcommand{\bfxp}{\mathbf{x}_p}

\newcommand{\bfw}{\mathbf{w}}

\newcommand{\bfI}{\mathbf{I}}
\newcommand{\bft}{\mathbf{t}}

\newcommand{\bff}{\mathbf{f}}

\newcommand{\bftheta}{{\boldsymbol{\theta}}}

\newcommand{\bfs}{\mathbf{s}}

\newcommand{\pd}{p_{\mathrm{data}}}

\newcommand{\mbb}[1]{\mathbb{#1}}

\newcommand{\ud}{\mathrm{d}}

\newcommand{\norm}[1]{\left\lVert#1\right\rVert}

\usepackage{mdframed}

\DeclareMathOperator*{\argmin}{arg\,min}

\title{Quantifying Sample Anonymity in Score-Based Generative Models with Adversarial Fingerprinting}

\author{%
  Mischa Dombrowski \textsuperscript{1}, Bernhard Kainz \textsuperscript{1,2} \\
  \and
  \textsuperscript{\rm 1} Friedrich-Alexander-Universität Erlangen-Nürnberg, DE \\
  \textsuperscript{\rm 2} Department of Computing, Imperial College London, London, UK \\
  \and
  \tt\small{mischa.dombrowski@fau.de}\\
}

\begin{document}

\maketitle


\begin{abstract}
Recent advances in score-based generative models have led to a huge spike in the development of downstream applications using generative models ranging from data augmentation over image and video generation to anomaly detection. 
Despite publicly available trained models, their potential to be used for privacy preserving data sharing has not been fully explored yet. 
Training diffusion models on private data and disseminating the models and weights rather than the raw dataset paves the way for innovative large-scale data-sharing strategies, particularly in healthcare, where safeguarding patients' personal health information is paramount.
However, publishing such models without individual consent of, e.g., the patients from whom the data was acquired,  necessitates guarantees that identifiable training samples will never be reproduced, thus protecting personal health data and satisfying  the requirements of policymakers and regulatory bodies. 
This paper introduces a method for estimating the upper bound of the probability of reproducing identifiable training images during the sampling process. This is achieved by designing an adversarial approach that searches for anatomic fingerprints, such as medical devices or dermal art, which could potentially be employed to re-identify training images.
Our method harnesses the learned score-based model to estimate the probability of the entire subspace of the score function that may be utilized for one-to-one reproduction of training samples. To validate our estimates, we generate anomalies containing a fingerprint and investigate whether generated samples from trained generative models can be uniquely mapped to the original training samples.
Overall our results show that privacy-breaching images are reproduced at sampling time if the models were trained without care. 
\end{abstract}

\section{Introduction}
Maintaining privacy and anonymity is of utmost importance when working with personal identifiable information, especially if 
data sharing has not been individually consented and thus cannot be 
shared with other institutions~\cite{jin2019review}. The potential of privacy preserving consolidating of private datasets would be significant and could potentially solve many problems, including racial bias \cite{larrazabal2020gender} and the difficulty of applying techniques such as robust domain adaptation \cite{wang2022metateacher}.  
Recent advances in generative modeling, \emph{e.g.}, effective diffusion models \cite{song2020denoising,dhariwal2021diffusion,rombach2022high,ruiz2022dreambooth}, enabled the possibility of model sharing~\cite{pinaya2022brain}. 
However, it remains unclear to what extent a shared model reproduces training samples  and whether or not this raises privacy concerns.

In general the idea of our research is to take a dataset $D$ of samples from the image distribution $p_{data}(\mathbf{x})$.
Then the goal is to train a generative model $s$, which learns only on private data. 
Direct privacy breaches would occur if the generative model exhibits a non-zero probability for memorizing and reproducing samples from the training set. 

Guarantees that such privacy breaches will not occur would ultimately allow to train models based on proprietary data and share the models instead of the underlying data sets. Healthcare providers would be able to share complex patient information like medical images on a population basis instead of needing to obtain individual consent from patients, which is often infeasible. Guarantees that no personal identifiable information is shared would furthermore pave the way to population studies on a significantly larger scale than currently possible and allow to investigate bias and fairness of downstream applications on anonymous distribution models of sub-populations. 

However, currently trained and published models can be prompted to reproduce training data at sampling time. \citet{somepalli2023diffusion} have observed that diffusion models are able to reproduce training samples and \citet{carlini2023extracting} have even shown how to retrieve faces of humans from training data, which raises serious privacy concerns. Other generative models are directly trained for memorization of training samples~\cite{cong2020gan}. 
We propose a scenario with an adversarial that has some prior information about a training sample and would therefore be able to filter out the image based on this information. In medical imaging this could be any medical device, a skin tattoo, an implant, or heart monitor; any detectable image with visual features that are previously known. 
Then an attacker could generate enough samples and filter images until one of the generated samples contains this feature. 
If the learned marginal distribution of the generative model that contains this feature is slim, then all images generated with it will raise privacy concerns. 
We will refer to such identifiable features as fingerprints. 
To estimate the probability of reproducing fingerprints, we propose to use synthetic anatomical fingerprints (SAF), which can be controlled directly through synthetic manipulations of the training dataset and reliably detected in the sampling dataset.

Our main contributions are: 
\begin{itemize}
    \item We formulate a realistic scenario in which unconditional generative models  exhibit privacy problems due to the potential of training samples being reproduced.
    \item We propose a mathematical method for finding the upper bound for the probability of generating sensitive data from which we derive an easily computable indicator. 
    \item We evaluate this indicator by computing it for different datasets and show evidence for its effectiveness.
\end{itemize}

\section{Background}
Consider $D$ containing samples from the real image distribution $p_{data}(\mathbf{x})$.
In general, highly effective generative methods like diffusion models~\cite{rombach2022high} work by modeling different levels of perturbation $p_{\sigma}(\tilde{\bfx}) \coloneqq \int p_{data}(\bfx)p_{\sigma}(\tilde{\bfx} \mid \bfx)\ud\bfx$ of the real data distribution using a noising function defined by $p_{\sigma}(\tilde{\bfx} \mid \bfx) \coloneqq \mathcal{N}(\tilde{\bfx}; \bfx, \sigma^2 \bfI)$. 
In this case $\sigma$ defines the strength of the perturbation, split into N steps $\sigma_{1}, \dots, \sigma_{N}$. 
The assumption is that $p_{\sigma_1}(\tilde{\bfx} \mid \bfx) \sim p_{data}(\mathbf{x})$ and $p_{\sigma_N}(\tilde{\bfx} \mid \bfx) \sim  \mathcal{N}(\bfx; \textbf{0}, \sigma_N^2\bfI)$

Then we can define the optimization as a score matching objective by training a model $\bfs_{\bftheta}(\bfx, \sigma)$ to predict the score function $\nabla_\bfx \log p_{\sigma}(\bfx)$ of the noise level $\sigma \in \{\sigma_i\}_{i=1}^{N}$. 

\begin{align}
\bftheta^* &= \argmin_\bftheta
   \sum_{i=1}^{N} \sigma_i^2  \mbb{E}_{\pd(\bfx)}\mbb{E}_{p_{\sigma_i}(\tilde{\bfx} \mid \bfx)}\big[\norm{ \bfs_\bftheta(\tilde{\bfx}, \sigma_i) - \nabla_{\tilde{\bfx}} \log p_{\sigma_i}(\tilde{\bfx} \mid \bfx)}_2^2 \big]. \label{eqn:ncsn_obj}
\end{align}

For sampling, this process can be reversed, for example, using Markov chain Monte Carlo estimation methods following~\citet{song2019generative}. 
\citet{song2020score} extended this approach to a continuous formulation by redefining the diffusion process as a process given by a stochastic differential equation (SDE): 

\begin{align}
    \ud \bfx = \bff(\bfx, t) \ud t + {g}(t) \ud \bfw \label{eqn:forward_sde},  
\end{align}

and training a dense model on predicting the score function for different time steps t, where $\bfw$ models the standard Wiener process, $\bff$ the drift function of $\bfx(t)$, that models the data distribution, and $\bfx(t)$ the drift coefficient. 
Therefore, the continuous formulation of the noising process, denoted by $p_t(\bfx)$ and $p_{st}(\bfx(t) \mid \bfx(s))$, is used to characterize the transition kernel from $\bfx(s)$ to $\bfx(t)$, where $0 \leq s < t \leq T$.

\citet{anderson1982reverse} show that the reverse of this diffusion process is also a diffusion process. The backward formulation is  
\begin{align}
    \ud \bfx = [\bff(\bfx, t) - g(t)^2  \nabla_{\bfx}  \log p_t(\bfx)] \ud t + g(t) \ud \bar{\bfw},\label{eqn:backward_sde}
\end{align}

which also extends the formulation of the discrete training objecting to the continuous objective:  

\begin{align}
   \bftheta^* = \argmin_\bftheta 
   \mbb{E}_{t}\Big\{\lambda(t) \mbb{E}_{\bfx(0)}\mbb{E}_{\bfx(t) \mid \bfx(0) }
   \big[\norm{\bfs_\bftheta(\bfx(t), t) - \nabla_{\bfx(t)}\log p_{0t}(\bfx(t) \mid \bfx(0))}_2^2 \big]\Big\}%
   . \label{eqn:training}
\end{align}

In Eq.~\ref{eqn:training}, $\lambda(t): [0, T] \to \mbb{R}_{>0}$ is a weighting function, often neglected in practice. 
\citet{song2020score} show that the reverse diffusion process of the SDE can be modeled as a deterministic process as the marginal probabilities can be modeled deterministically in terms of the score function. As a result, the problem simplifies to an ordinary differential equation: 

\begin{align}
    \ud \bfx = \Big[\bff(\bfx, t) - \frac{1}{2} g(t)^2\nabla_\bfx \log p_t(\bfx)\Big] \ud t, \label{eqn:deterministic_flow}
\end{align}

and can therefore be solved using any black box numerical solver such as the explicit Runge-Kutta method. This means that we can perform exact likelihood computation, which is typically done in literature, to estimate how likely the generation of test sample, \emph{e.g.} images, is. This means that low negative log-likelihood (NLL) is desirable. In our case, we want to estimate the likelihood of reproducing training samples at test time. Ideally, this probability would be zero or very close to zero.

\section{Method}
Typically, NLL measures how likely generating test samples at training time is. To use it to evaluate the memorization of training data, we compute the NLL of the training dataset. 
A limitation of using NLL is that it only computes the likelihood of the exact sample to be reproduced at sampling time and therefore is insufficient for giving estimates of the likelihood of generating samples that raise privacy concerns.
We can compute the likelihood of the exact sample but this does not mean that the images in the immediate neighborhood are not leading to privacy issues. 
We assume that all images of the probability distribution are within a certain 
distance to the real image. 
As a result, we propose to estimate the upper bound of the likelihood of reproducing samples from the entire subspace that belongs to the class of private samples. 
First, we define the sample $\bfxp$ that we consider to be a potential privacy breach and augment this sample by adding a synthetic anatomic fingerprint (SAF) to it. 
This SAF is used to identify the sample, which raises privacy concerns. Then we repeatedly apply the diffusion and reverse diffusion process and check when the predicted sample starts to diverge to a different image.

\subsection{Estimation Method}
\label{sec:estimation_method}
Let $p_s(\bfxp)$ define the likelihood of the model $\bfs$ to reproduce  the private sample $\bfxp$  at test time. Following \cref{eqn:deterministic_flow}, we can  compute the likelihood of this exact sample. However, this does not account for the fact that images in the immediate neighborhood, like slightly noisy versions of $\bfxp$, are not anonymous. Consequently, we are interested in computing $q(p)$, which is defined as the likelihood of reproducing any sample that is similar enough to the target image that it raises privacy concerns: 

\begin{align}
   q(p) = \int_{\Omega_p} p_s(\bfx)\ud\bfx \label{eq:pofp},
\end{align}

where $\Omega_p$ is defined as the region of the image $\bfxp$ that is private. We determine this region by training a classifier tasked with detecting whether the image belongs to the image class, as explained in \cref{sec:boundary_computation}.
To search through the image manifold, we make use of the reverse diffusion process centered around the SAF image $\bfxp$ defined as $p_{t,b}  \coloneqq p(\bfx_t \mid \bfxp) =  \mathcal{N}(\tilde{\bfx}; \bfxp, \sigma_t^2 \bfI)$ for $\bfx(s)$ to $\bfx(t)$, where $0 \leq t \leq T$. 
We can employ the diffusion process centered around this image to sample from the neighborhood and then use the learned reverse diffusion process to generate noisy samples $\bfx_{t,p}$. Then we can use this as starting image for the reverse diffusion process to sample $\bfx_{t,p}'$:

\begin{align}
   q(p) = \int_{\Omega_{p}} p_s(\bfx)\ud\bfx \approx \int_0^{t'} p_s(\bfx_{t,p}) \ud\bft = \int_0^{t'}\mbb{E}_{p(\bfx_{t,p})}\big[    p(\bfx'_{t,p})\big]\ud\bft .
\end{align}

Technically, we could employ exact likelihood computation to estimate $q(p)$ but this would require integrating over the continuous image-conditioned diffusion process, which would be intractable in practice. Therefore, we propose to approach and estimate this integral by computing the Riemann sum of this integral and give an upper bound estimate for it using the upper Darboux sum: 

\begin{align}
\int_0^{t'}\mbb{E}_{p(\bfx_{t,p})}\big[    p(\bfx'_{t,p})\big]\ud\bft = \sum_{t} (\sigma_t - \sigma_{t-1}) \mbb{E}_{p(\bfx_{t,p})}\big[p(\bfx'_{t,p})\big] \leq \\ \sum_{i=0}^{t'} \sup_{t \in\left[t_i, t_{i+1}\right]} (\sigma_{t_{i+1}} - \sigma_{{t_i}})\mbb{E}_{p(\bfx_{t,p})}\big[p(\bfx'_{t,p})\big], \label{eq:full_equation_estimate} 
\end{align}

which approaches the real value for steps that are small enough. We can compute this value by using $\bfxp$ as a query image and then estimating the expectation by performing Monte-Carlo sampling but this would still require a lot of time due to the computational complexity of exact likelihood estimation.

\subsection{Method intuition}
\begin{figure}
      \centering
      \fbox{
      \includegraphics[width=0.95\linewidth]{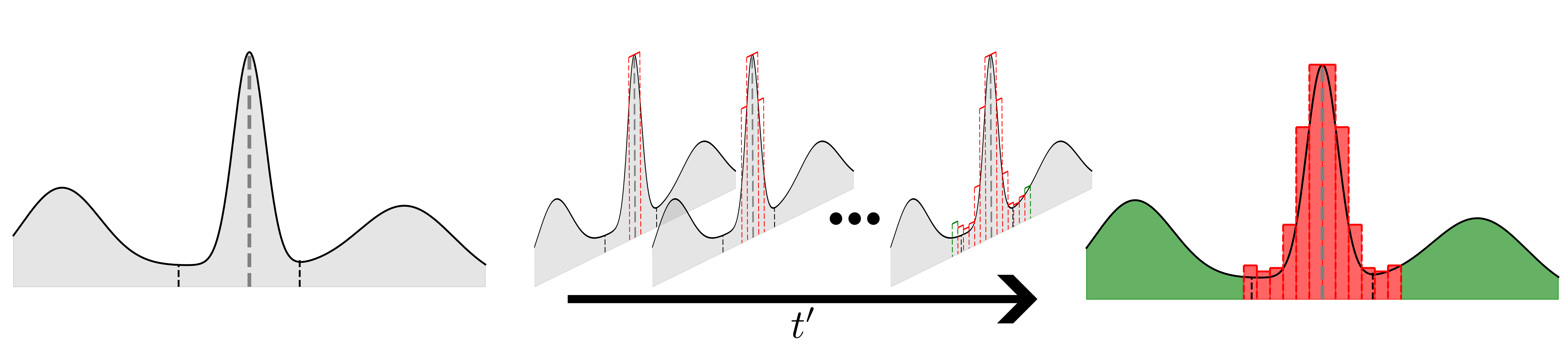}
      }
    \caption{Illustration of our estimation method in 1D. The grey line denotes the query image $\bfxp$. The estimation method iteratively increases the search space in the latent space of the generative model. The green area corresponds to image regions resulting in non-privacy concerning generated samples, while the red area is considered critical. }
    \label{fig:illustrationofmethodin1D}
\end{figure}

Intuitively, we model the image space using the learned distribution of the score function $\nabla_{\tilde{\bfx}} \log p_{\sigma_i}(\tilde{\bfx} \mid \bfx)$ by reversing the diffusion process and checking when the model starts to ``break out'' by generating images classified as different samples. 
For large $t$, the learned marginals $p(\bfx, t)$ span the entire image space. Importantly, by definition of the diffusion process, the distribution approaches the same distribution as the sampling distribution of the diffusion process if $\sigma_t$ gets large enough $p_{\sigma_N}(\tilde{\bfx} \mid \bfxp) \sim  \mathcal{N}(\bfx; \textbf{0}, \sigma_N^2\bfI)$. However, for lower $t$ the model has learned that the distribution collapses towards a single training image $\bfxp$. Essentially, it has modeled part of the subspace as a delta distribution around $\bfxp$. 
We want to find out how far back in the diffusion process we have to go for the model to start to produce different images. 
The boundary $\Omega_p$ is then defined as all images that would collapse towards this training image and estimated using the classifiers. 
\cref{fig:illustrationofmethodin1D} illustrates this process in one dimension. 
Note that this is different from simply defining a variance that is large enough for the classifiers to fail, as $s_{\theta}(\bfxp, \sigma_t)$ was trained to revert this noise.

\subsection{Synthtetic Anatomic Fingerprint}
Let $\Tilde{\mathit{D}}$ be our real dataset of size $N$ without any privacy concerns due to the lack of any identifiable information. Then we synthetically generate a dataset $\mathit{D}$ which contains a single sample with a fingerprint. 
Importantly, we remove the non-augmented version of that sample from the  $\mathit{D}_p$.
In practice, this can be any kind of fingerprint that appears only once in the entire training dataset. To ease the training of identification classifiers, we choose a grey constant circle as shown in \ref{fig:visualabstract}. 
Therefore, the SAF sample $\bfxp$ is defined as an augmented version of a real sample: 
\begin{align}
    \bfxp = \bfx_i * (1 - L_{p}) + \bfx_{SAF} * L_{p}
\end{align}
where $i$ is a randomly drawn sample from $D$. The location of the SAF determined by $L_{p}$ is randomly chosen to lie entirely within the boundary of the image. 
Then we train a score-based model $\bfs_\bftheta(\bfx, \sigma_t)$ on the augmented dataset $\mathit{D}_p$. 
To quantify whether or not the trained model is privacy concerning, we define an adversarial attacker that knows of the fingerprint $\bfx_{SAF}$ and that we can train on detecting this fingerprint. We will refer to this classifier as $\cp$. 
The second classifier $\cid$ is trained on $\mathit{D}$ in a one-versus-all approach to classifying the image's identity. 
We assume that private information is given away when this classifier correctly detects the generated sample. 
Importantly, we train $\cid$ with random masking using the same circular patches that were used to generate $L_{p}$. Therefore we can use $\cp$ to filter out all images that contain the SAF and then determine whether or not this sample raises privacy concerns by computing the prediction for $c_{id}(\bfx')$ generated samples from the generative model $s_{\theta}$
\subsection{Boundary Computation}
\label{sec:boundary_computation}
To give an estimate for $q(p)$ we observe that it only depends on the likelihood $p(\bfxp)$ and $t'$, which is supposed to capture the entire region of $\Omega_p$. 
Therefore, we use $\bfxp'$ as input to the classifiers and define $\Omega_p$ as the region where both classifiers give a positive prediction. 
Since exact likelihood computation and the terms for the variance derived in \cref{eq:full_equation_estimate} reach computationally infeasible value ranges, we can use $t'$ as an estimate of how unlikely it is to generate critical values from the model.

\label{sec:estimation_algorithm}
\begin{algorithm}[t!]
\caption{Upper bound likelihood estimation algorithm}
\label{alg:CKB}
\DontPrintSemicolon
\KwIn{$M$, $s_\theta(\bfx, t)$, $c_{SAF}(\bfx)$, $c_{ID}(\bfx)$, $\bfxp$}
\KwResult{$t'$}
\BlankLine
\For{$t\!=\!1,\dots,0$}{
\For{$m\!=\!1,\dots,M$}{
    $\bfx_{t,p}$ \!= $p(\bfx_t \mid \bfx_p)$\;
    \For{$\tilde{t}\!=\!t,\dots,0$}{
        $\bfx_{t,p}'$ \!= $s_\theta(\bfx_{t,p}'$, $\tilde{t}$)\;
    }
    $\bfxp'$ = $\bfx_{t,p}'$\;
    \If{$c_{SAF}(\bfx)$ is True \textbf{and} $c_{ID}(\bfx)$ is True}{
        \textbf{return} t\;
    }

}
}
\end{algorithm}

\begin{figure}
      \centering
      \fbox{
      \includegraphics[width=0.95\linewidth]{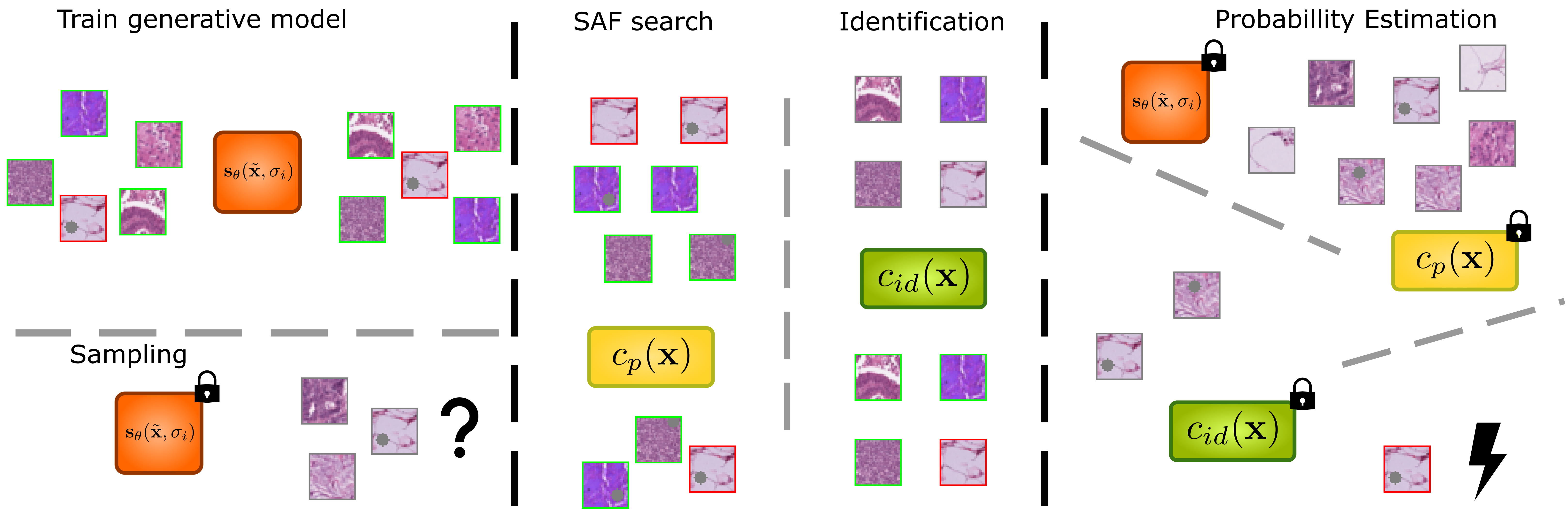}
      }
    \caption{Visual abstract of our evaluation method for privacy problems. Uncritical samples are shown in green and critical samples in red. The SAF is injected into the training pipeline of the generative mode. SAF search and identification is done using supervised training with samples from the training set. Finally, we filter generated image looking for samples that contain this SAF and check if we can identify the image.}
    \label{fig:visualabstract}
\end{figure}

\cref{alg:CKB} describes the computation of $t'$. We can freely choose $M$ as parameter and trade-off accuracy for computation time.
Given $\bfxp$ we define $q_M(p|x_{t,p})$ as the estimate of staying within the boundary of $\Omega_p$ for a given diffusion step $t$. Then we define $t' \coloneqq \text{max}(\mathbb{T}), \mathbb{T} \coloneqq \{ \forall t \colon q_M(p|x_{t,p}) > 0 \} $. The full pipeline is illustrated in \cref{fig:visualabstract}


%




\section{Related Work}
Generative models have disrupted various fields by generating new data instances from the same distribution as the input data. These models include Variational Autoencoders (VAEs)~\cite{kingma2013auto}, Generative Adversarial Networks (GANs)~\cite{goodfellow2014generative}, and more recently, Generative Diffusion Models (GDMs). 
Diffusion models can be categorized as score-based generative models~\cite{song2019generative,song2020score} and models that invert a fixed noise injection process~\cite{sohl2015deep,ho2020denoising}. In this work we focus on score-based generative models. 

Evaluating data privacy in machine learning has been a longstanding concern~\cite{dwork2006calibrating,abadi2016deep,van2021memorization}. 
Research on integrating privacy-preserving mechanisms in generative models is still in its infancy. \cite{xie2018differentially} proposed a method to make GANs deferentially private by modifying the training algorithm. \cite{jiang2022dp} applied differential privacy to VAEs, showcasing the possibility of explicitly integrating privacy preservation into generative models.

Despite the progress in privacy-preserving generative models, little work has been done on evaluating inherent privacy preservation in diffusion models and providing privacy guaranteed dependant on the training regime. To the best of our knowledge, our work is the first to investigate natural privacy preservation in generative diffusion models, contributing to the ongoing discussion of privacy in machine learning.

\section{Experiments}
\subsection{Dataset}
For our experiments we use MedMNISTv2 \cite{DBLP:journals/corr/abs-2110-14795}. This dataset consists of a combination of multiple downsampled $28 \times 28$ images from different modalities. Some of them are multichannel, while others are single-channel. For single-channel images we repeat the channel dimension three times. 
For our main experiments we choose PathMNIST, due to the high amount of samples available for that dataset. Furthermore, we experiment with an a-priori selected set of modalities from this dataset which ranges through multiple sizes and multiple  channels of the dataset. 
\subsection{Models}
The classifiers are randomly initialized ResNet50 \cite{resnet50} architectures. 
To maximize robustness we employ AugMix \cite{hendrycks2020augmix} and in the case of $\cid$ we furthermore inject random Gaussian noise into the training images to increase the robustness towards possible artifacts from the diffusion process. 
Furthermore, we randomly mask out patches of the same shape as the SAF to reduce the effect of SAF on the prediction.  
The training and sampling of the score-model follows the implementation of \cite{song2020score} with sub-VP SDE sampler due to their reported good performance on exact likelihood computation \cite{song2020score} with a custom U-Net architecture based on \cite{von-platen-etal-2022-diffusers}. 
Training $s_{\theta}$ is done on a single A100 GPU and takes roughly eleven hours. The classifiers are trained until convergence with a patience of 20 epochs in less than one hour. Exhaustive search for $t'$, which is done by computing $q_{M=16}(p|x_{t,p})$ for all $t \in {0, \dots, 1}$, takes four hours. 

\subsection{Reverse Diffusion Process}
\label{sec:rev_diff_proce}

\begin{figure}
      \centering
      \fbox{
      \includegraphics[width=0.95\linewidth]{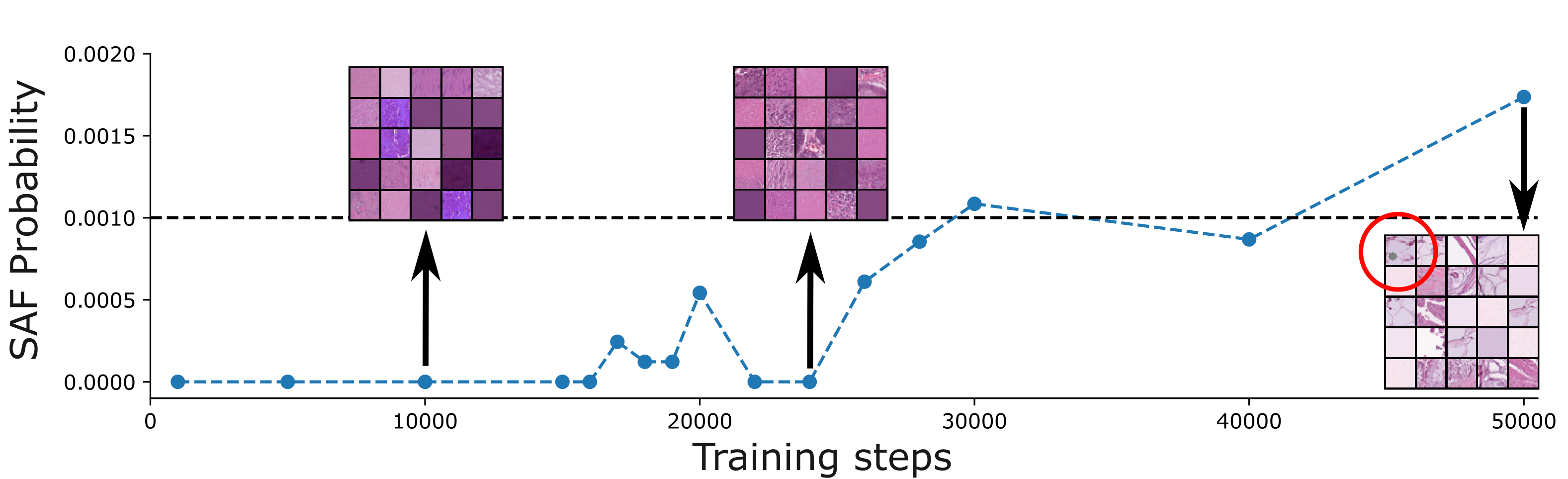}
      }
    \caption{Influence of training length on generative and memorization properties. A positively classified sample can be seen in the top-left corner of the rightmost image.}
    \label{fig:epoch_vs_probablity}
\end{figure}

\begin{figure}
	\centering
	\fbox{
		\includegraphics[width=0.95\linewidth]{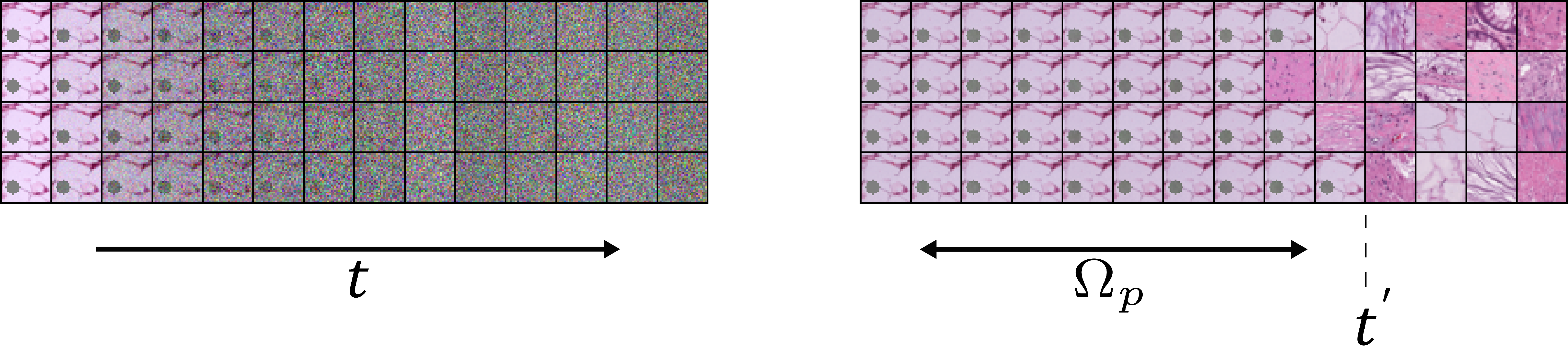}
	}
	\caption{Illustration of the reverse diffusion process. Left shows query images  $\bfx_{t,p}$ for $t \in \left[0, 0.7\right]$. Right shows the resulting sample.}
	\label{fig:reversediffusion}
\end{figure}

First, we experiment with the influence of the training length on $|p|$ by sampling 10000 images from a model trained on $|N_{D}| = 1000$ and show the results in \cref{fig:reversediffusion}.
For the first 14000 steps, the model only learns high-frequency attributes of the data. The visual quality is low and therefore also the probability of reproducing $\bfxp$ Around 20000 the quality of the generated samples improves visually, but also the number of memorized training samples. At this point, the model already starts to accurately reproduce $\bfxp$ at sampling time. Every detected sample is visually indistinguishable from the training image. The MAE even goes down to  $1 \times 10^{-4}$. Based on these observations, we continue our investigations with a fixed training length of 30000 steps. 

\begin{figure}
      \centering
      \fbox{
      \includegraphics[width=0.95\linewidth]{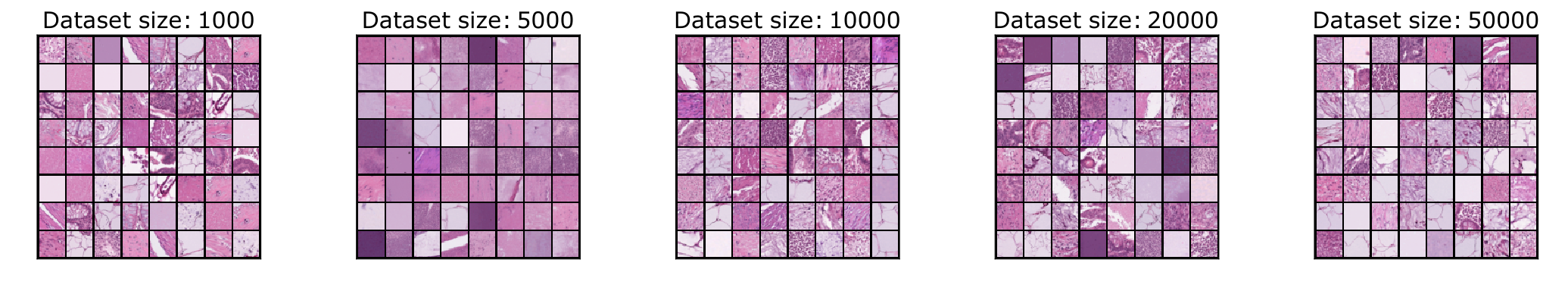}
      }
      \caption{Representative samples from trained models on different dataset sizes $|N_{D}|$.}
      \label{fig:datasetsizeexperiment}
\end{figure}

Next we want to investigate the influence of the size of the dataset on its memorization capabilities. Therefore we train models on different $|N_D|$sample 150000 images for every model and at testing the probability of reproducing our sample at test time. 
We do this by defining the null-hypothesis $H_{0}$ that the probability of sampling $\bfxp$ is equal to $1/N_{D}$. $H_{1}$ claims that the probability is lower. Therefore, we sample 150000 images for every trained model with dataset size $|N_{D}| \in \left\{ 1000, 5000, 10000, 20000, 50000 \right\}$. The results are shown in \cref{tab:datasetsizetable}
It can be seen that the model only learned to reproduce samples with the SAF when the dataset size was comparably low.
For $|N_{D}| = 1000$ the model was surprisingly close to the expected value, indicating that the size of the data is too small relative to the available parameter space and the model memorizes them as discrete distribution of $1000$ unrelated images. 
Every other model produces very few positive predictions from the classifier all of which turn out to be false positives. 
The combined prediction $q \coloneqq \cid^+ \cap \cp^+$ is only positive for the smallest dataset. 
All the larger models don't have any positive samples in their dataset. The p-value for this is smaller than 5\% in all cases, meaning that we can reject the null-hypothesis and assume that the probability of $\bfxp$ is smaller.
Next we look at the samples of different sizes and show them in \cref{fig:datasetsizeexperiment}. Initial observation suggest that image quality drops for medium-sized datasets. 
However, upon closer inspection we see that the smallest model simply learns to reproduce training data, which can be seen by the fact that some images appear multiple times. 
This confirms our observation that the model learned the training distribution in the form a discrete set of 1000 images but never learned to generalize. 
In the context of data-sharing this would mean that the model is simply a way of saving training data but would still raise privacy concerns.
The model trained on 5000 images seems to lie in between generalizing and memorizing the learned distribution but the size of dataset was not large enough to learn a meaningful representation. The result looks like it learned low frequency information such as color or larger structure, but the images are lacking detail. 

\begin{table}
	\caption{Number of positive predictions of the classifiers for models trained on different dataset size on 150000 images. All models use the same classifiers.}
	\label{tab:datasetsizetable}
	\centering
	\begin{tabular}{llllll}
		\toprule
		$|N_{D}|$& 1000  & 5000   & 10000 & 20000 & 50000 \\
		\midrule
		$\mbb{E}\left[|q|\right]$ & 150 & 30 & 15& 7.5& 3\\
		\midrule
		$|\cp^+|$ & 151 & 0 & 0& 1& 1 \\
		$|\cid^+|$& 151 & 0 & 3 & 3 & 4 \\
		$|q|$ & 151 & 0 & 0&0&0 \\
		\bottomrule
	\end{tabular}
\end{table}

\begin{figure}
      \centering
      \fbox{
      \includegraphics[width=0.95\linewidth]{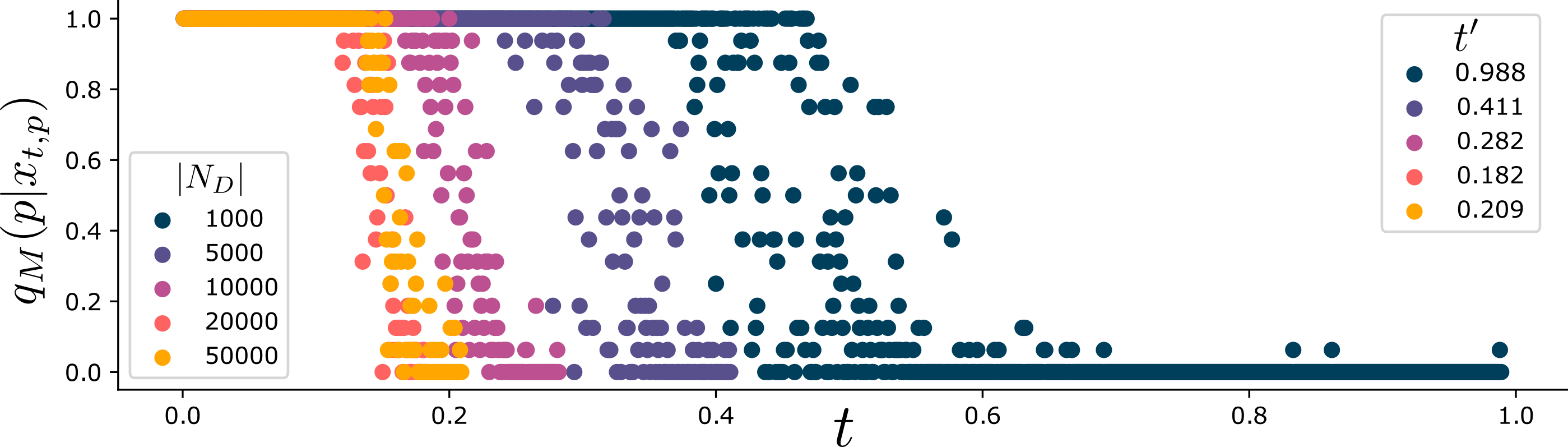}
      }
    \caption{Likelihood of producing $\bfxp$ at sampling time as a function of $t$ for $t \in \{0, \dots, 0.5\}$ and $M = 16$. We stop plotting probabilities after $t'$.
    Due to the high observed probabilities of the $N_D = 1000$ model, we also compute and plot the probabilities for higher t.}
    \label{fig:likelihood_t_dash}
\end{figure}

Now we can use our proposed estimation method from \cref{alg:CKB} to compute $t'$ for all datasets with $M=16$. The results are shown in  \cref{fig:likelihood_t_dash}. Clearly, the probability for generating samples $q_M(p|x_{t,p})$  decreases with increasing t. More importantly, the threshold at what point the probability drops, is higher for smaller $|N_D|$, which means $t'$ is indeed an important indicator for $q(p)$. Additionally, these results show that sharing the model with $|N_D| = 5000$ would raise more privacy issues as other modes, as the indicator suggests that the probability for a sample being generated at inference time is high. 

Finally we validate our results by looking at different datasets in \cref{tab:SAFTrainingResults}. 
The results confirm our observations of a high amount of memorization in models with small dataset sizes close to the expected value. 
There is once again a turning point at around 5000 images where samples are no longer memorized. We can also confirm this gap by comparing the FID by calculating it on the training and test dataset. The drop is large in all cases where training samples are memorized. 
However, FID fails to measure the extent of this effect. PneumoniaMNIST has a larger drop in performance than RetinaMNIST but barely any memorized samples. 
Our proposed indicator $t'$ on the other hand captures this observation. 
Furthermore, it is also lower for the BreastMNIST dataset, which according to the high difference between $\mbb{E}(|q|)$ and $|q|$, did not collapse as strongly towards only reproducing $\bfxp$.

\begin{table}
  \caption{Training results for different MedMNIST datasets. We report test accuracy for the SAF classifier but only training accuracy for the ID classifier as identification only makes sense if the sample was part of the training set. For the generative scores we use 50000 samples.}
  \centering
      \label{tab:SAFTrainingResults}
  \begin{tabular}{lrccccccc}
      \toprule
      \multicolumn{2}{c}{Description} & \multicolumn{2}{c}{SAF Classification} & \multicolumn{5}{c}{Data Synthesis} \\
       \cmidrule(r){1-2}\cmidrule(r){3-4}\cmidrule(r){5-9}
        Dataset  &$|N_D|$ & SAF (\%)& ID (\%) & $\text{FID}_{train}$ & $\text{FID}_{test}$ & $\mbb{E}(|q|)$ & $|q|$ & $t'$ \\
       \midrule
        RetinaMNIST    & 1080   & 100  & 99.6      & 5.9   & 19.7  & 46.3 & 52 & 0.998\\
        BloodMNIST     & 11959  &100   & 99.5     & 9.3   & 11.   & 4.2 & 0 & 0.241\\
        ChestMNIST     & 78468  &99.93 & 99.8  & 3.3   & 3.9   & 0.6 & 0 & 0.206\\
        PneumoniaMNIST & 4708   &100   & 99.8 & 9.5   & 28.4  & 10.6 & 2 & 0.719\\
        BreastMNIST    & 546    &100   & 98.7        & 9.2   & 62.6  & 91.6 & 57 & 0.886\\
        OrganSMNIST    & 13940  &99.47 & 99.8 & 19.6  & 19.7  & 3.6 & 0 & 0.582\\
     \bottomrule
  \end{tabular}
\end{table}

\section{Discussion}
We have shown that $t'$ is a useful indicator towards estimating $q(p)$ since it can be directly derived from it as shown in \cref{sec:estimation_method}. 
 Our results show that training and publishing trained models without care can lead to critical privacy breaches due to direct data-sharing. 
The results also suggest that SAFs are either memorized or ignored. This has important implications on the feasibility of using these models instead of direct data sharing as this impedes the ability to use the shared model for datasets with naturally occurring anomalies. These features are often crucial for medical applications but highly unlikely to be reproduced at sampling time, making detecting these features in downstream applications, such as anomaly detection, even harder.  
Computation of $t'$ does not necessarily require the existence of $c_{p}$, only that of $c_{id}$. Therefore, it can be applied as an indicator of overfitting in  diffusion-based generative models. 
The exhaustive search described in \cref{alg:CKB} could be approximated using improved search techniques such as binary search random subsampling of $t$ or using a reduced search range. 

\section{Limitations}~
Our experiments consider clear synthetic outliers that are not necessarily congruent to  the real image distribution. 
It would be interesting to see if the effect is different if the SAF is closer to the real image. 
However, the fact that they are visually distinguishable from everything else is necessary for the image to remain detectable and also for the assumption that they pose a privacy concern. 
Additionally, our experiments focus on a single way of training and sampling the models. 
However, current approaches such as~\cite{meng2023distillation} often use different samplers, training paradigms, or distillation methods. It remains to be shown whether or not these different approaches change the learned distribution of the underlying score model or if they only improve perceptual properties. 
Finally, due to the complexity of the problems and the high dimensionality, we do not compute a real estimate for the probability of the data to appear at sampling time but only an indicator. 
The indicator $t'$ has a high variance for large $t$ if $|N_D|$ is small due to the high stochasticity involved when sampling $q_M(p|x_{t,p})$. Therefore, results with $t'$ close to 1 are hard to compare against each other. But as we have shown, these are the cases in which the models raise a privacy concern and direct sampling of $\bfxp$ is possible according to \cref{tab:datasetsizetable}.

\section{Conclusion}
In this work we have described scenarios in which training score-based models on personal identifiable information like image data can lead to data-sharing issues. 
By defining an adversarial that has prior information about a visual property of the data, we showed that training and publishing these models without care can lead to critical privacy breaches. 
To illustrate this, we have derived an indicator for the likelihood of reproducing training samples at test time. 
The results show that generative models trained on small datasets or long training times should not be readily shared. 
Larger dataset sizes, on the other hand, lead to the model ignoring and never reproducing the detectable fingerprints. 
In the future, we will work on using $t'$ in an adversarial fashion to train models that are explicitly taught not to sample from these regions in the representation space.  

\noindent\textbf{Acknowledgements:} 
The authors gratefully acknowledge the scientific support and HPC resources provided by the Erlangen National High Performance Computing Center (NHR@FAU) of the Friedrich-Alexander-Universität Erlangen-Nürnberg (FAU) under the NHR project b143dc PatRo-MRI. NHR funding is provided by federal and Bavarian state authorities. NHR@FAU hardware is partially funded by the German Research Foundation (DFG) – 440719683.

\renewcommand{\thesection}{\Alph{section}}

%
%
%
%
%
%


\newpage
{
\small
\bibliographystyle{plainnat}
\bibliography{main}
}

\newpage

\appendix
\section{Model training details}
\begin{figure}
      \centering
      \fbox{
      \includegraphics[width=0.95\linewidth]{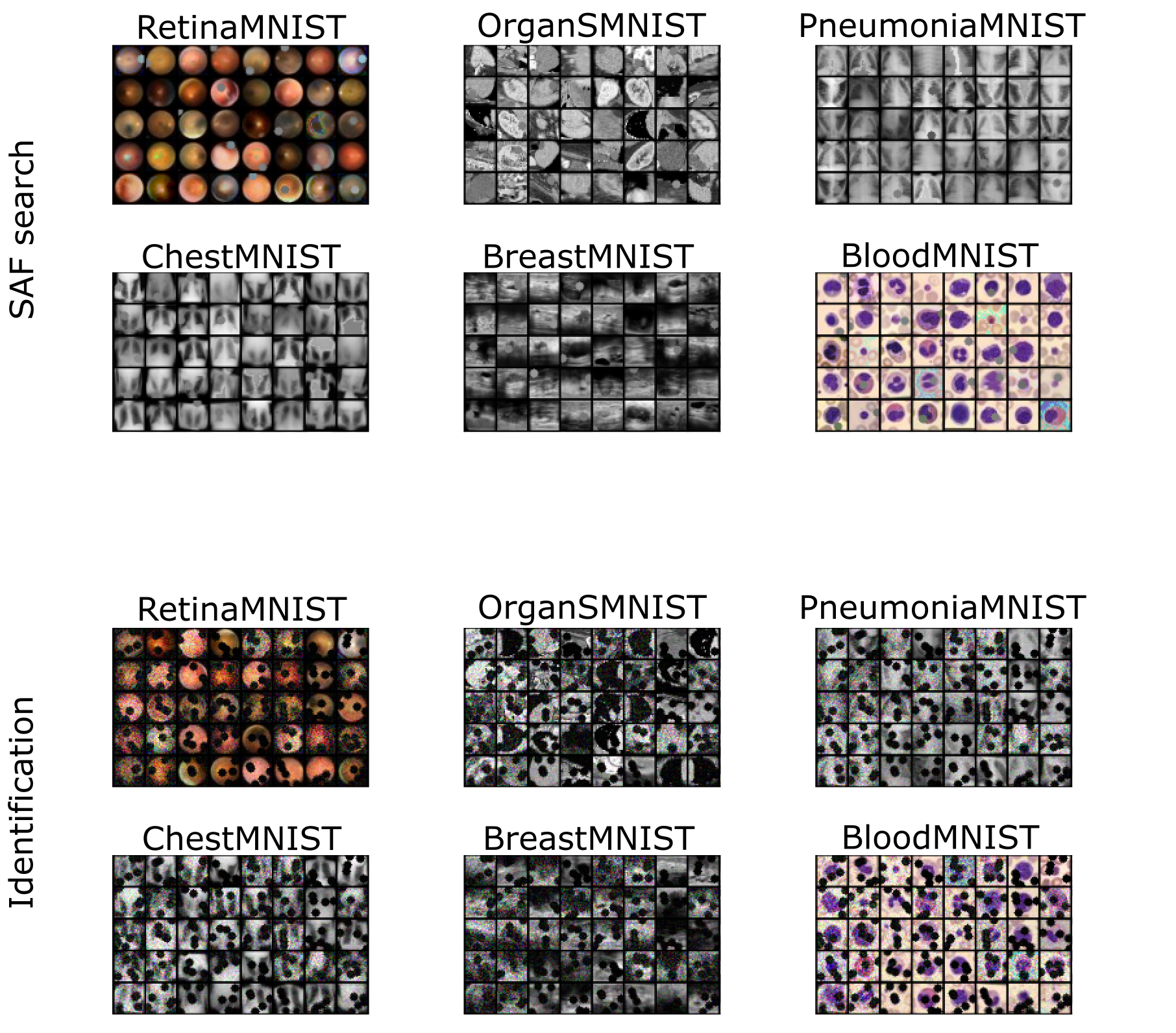}
      }
    \caption{Training image samples for $\cp$ and $\cid$}
    \label{fig:supp_training_images}
\end{figure}
\begin{figure}
      \centering
      \fbox{
      \includegraphics[width=0.95\linewidth]{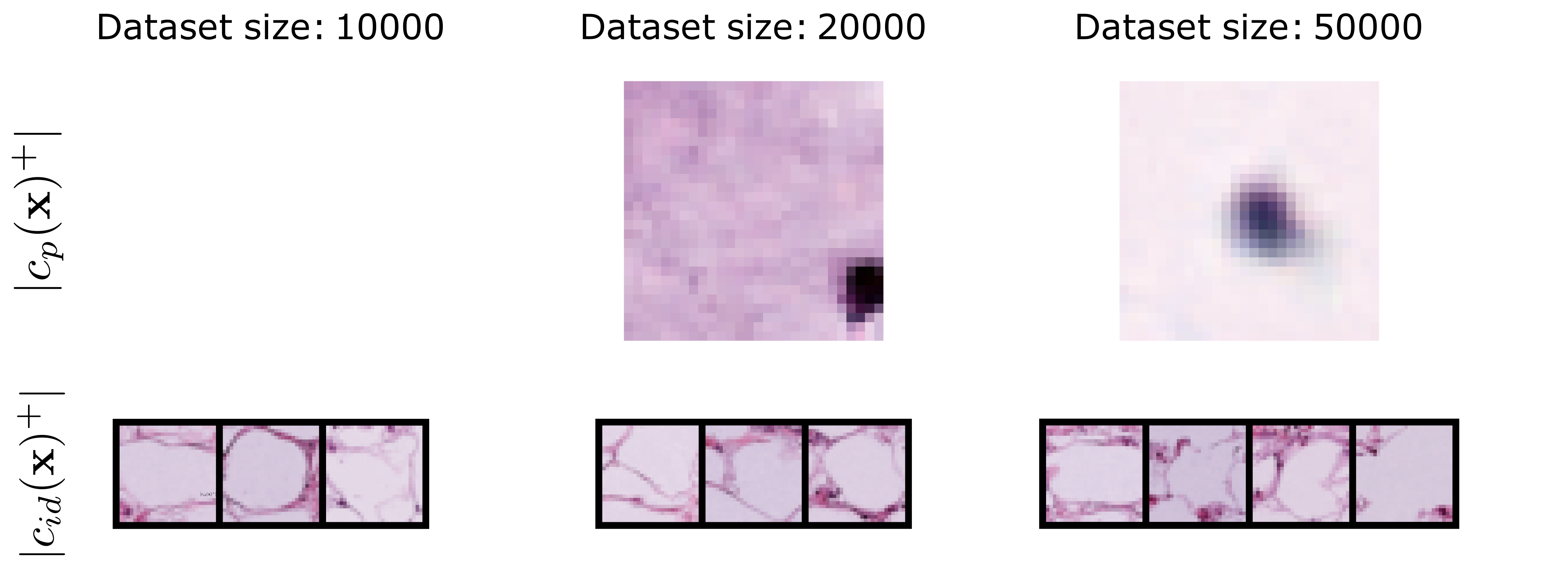}
      }
    \caption{All false positive predictions from the 750000 generated images. All misclassified images by one classifier were filtered and correctly classified by the other classifier.}
    \label{fig:false_positives}
\end{figure}

To further elaborate on the training details of $\cp$ and $\cid$, we show training samples for both classifiers in \cref{fig:supp_training_images}. 
Since both tasks are fairly easy binary classification tasks, we employed strong augmentation techniques to ensure that positively predicted samples from the classifiers are SAFs. 
We balanced the classification task for $\cid$ by adding SAFs to 50\% of the training images. For validation, we reduce this to 10\% to remain closer to the expected distribution. 
For $\cid$ we chose circular masking as training augmentation because we expected it might be necessary to mask out the SAF from the positive predictions of $\cp$. However, closer inspection of the predictions showed this was unnecessary (compare \cref{fig:false_positives}). Another reason is, that we do not want to confuse the model at inference time by showing it SAFs which are not part of the training data of $\cid$. 
The probability of $\bfxp$ appearing in the training dataset of $\cid$ is set to 10\% during training and 50\% during validation. 
The custom diffusion model architecture is based on the open-source implementation of a 2D U-Net\footnote{https://github.com/huggingface/diffusers}. Due to the $28 \times 28$ input images we are forced only to use the three outermost downsampling and upsampling layers.

\section{False positive predictions of $\cid$ and $\cp$}
The trained models only produce up to five false positives for 150000 generated images, as discussed in Section 5.3. The false positives for all $|N_{D}|$ are shown in \cref{fig:false_positives}. Both misclassified  samples from $\cid$ show great resemblance to the SAF by consisting of a circular monochrome patch. The misclassified  identification samples are really similar in terms of texture, color, and structure, although the differences to $\bfxp$ are distinct. None of the $\cid^+$ would lead to clear privacy issues in practice, which we successfully capture by computing $|q| = 0$ for these three models. 

\section{Detailed results for $t'$ on other datasets}
\begin{figure}
      \centering
      \fbox{
      \includegraphics[width=0.95\linewidth]{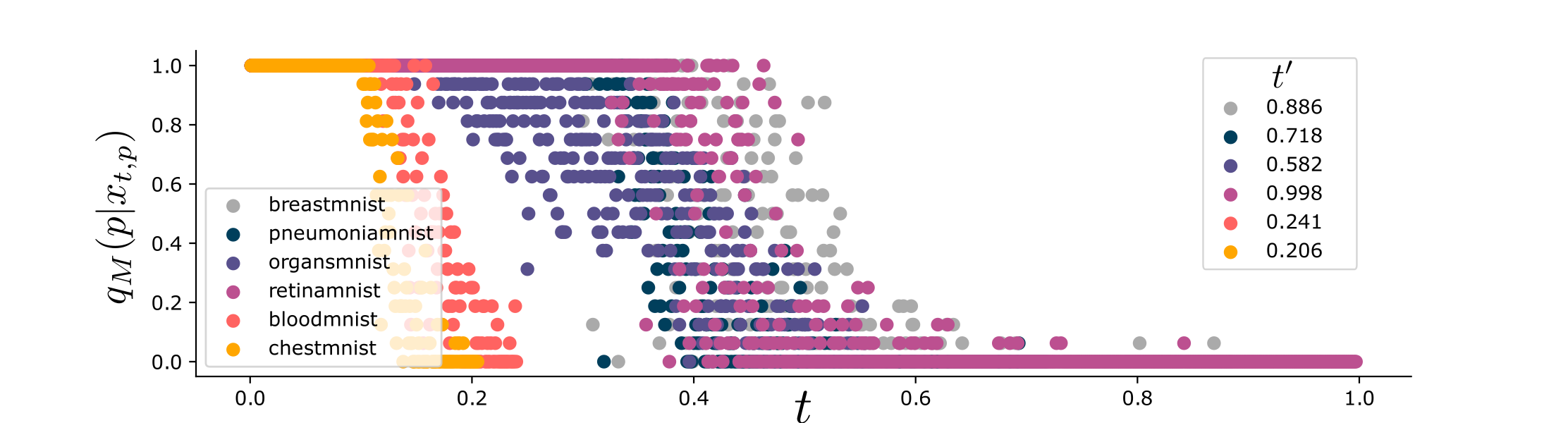}
      }
      
    \caption{Likelihood of producing $\bfxp$ at sampling time as a function of $t$ for $t \in \{0, \dots, 1\}$ and $M = 16$. We stop plotting probabilities after $t'$.}
    \label{fig:tdashmedmnistall}
\end{figure}

Next, we report the detailed results for other MedMNIST datasets. This time we perform an exhaustive search for $t'$ and visualize the results in \cref{fig:tdashmedmnistall}. The trained generative models exhibit the same behavior of starting a slow decline in the probability of reproducing training samples. The end of the decline can be estimated by computing $t'$.

\section{MAE of memorized training samples}
\begin{figure}
      \centering
      \fbox{
      \includegraphics[width=0.95\linewidth]{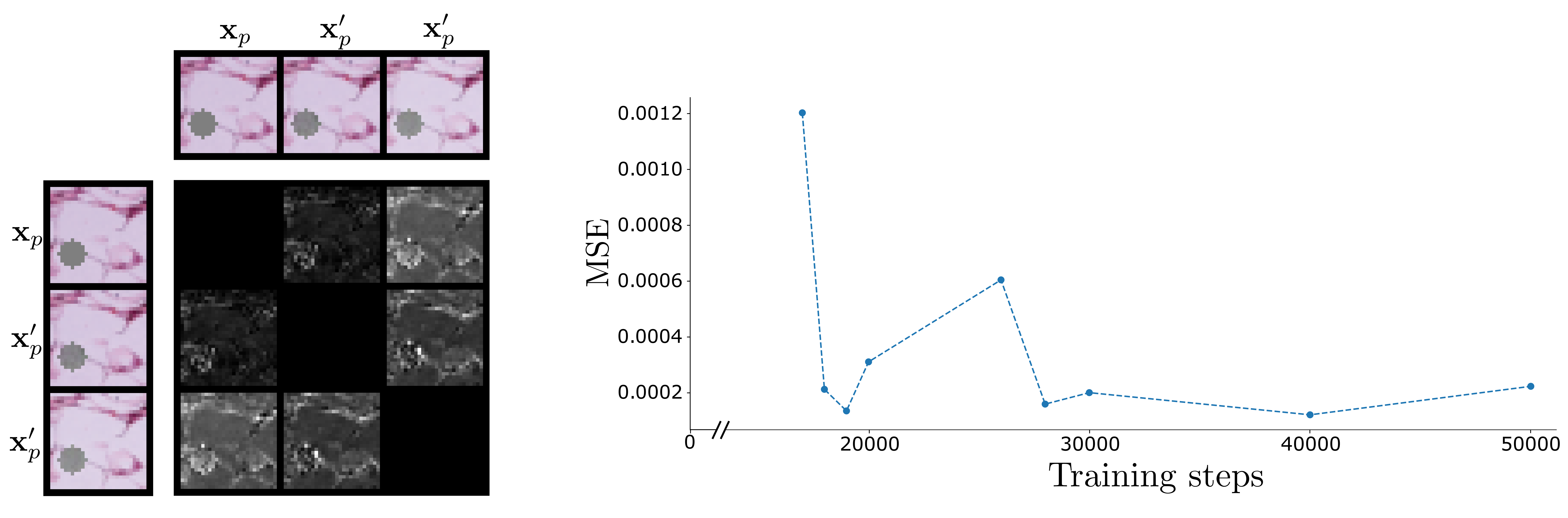}
      }
    \caption{The figure shows a grid-wise comparison of absolute pixel error between the training image $\bfxp$ and two sampled image $\bfxp'$ that raise privacy concerns (left) and the mean squared error (MSE) for an increasing amount of different training steps (right). $|N_D|$ is set to 1000. The samples on the left are from the model trained for 17000 steps.}
    \label{fig:suppmae}
\end{figure}
Our pipeline unveiled that training the score-based generative model for a long time on a small dataset leads to reproducing images at sampling time. We show this by applying our classification pipeline and filtering out all negative samples to get $q$. \cref{fig:suppmae} shows how much these samples are memorized. As can be seen, the sampled images $\bfxp'$ are barely distinguishable from the training image $\bfxp$. Interestingly, the mean squared error (MSE) between these images goes down rapidly but seems to stagnate after 19000 steps, at which point the reconstruction does not improve much, despite the observed higher memorization probability $q$ reported in Chapter 5.3. 
This suggests that overfitting occurs not only in the last reverse diffusion steps but also for higher $t$.

\medskip

\end{document}